\documentclass[11pt]{article}
\usepackage{coling2016}
\usepackage{times}
\usepackage{url}
\usepackage{latexsym}
\usepackage{times}
\usepackage{latexsym}
\usepackage{url}
\usepackage{graphicx}
\usepackage{epstopdf}
\usepackage{amsfonts}
\usepackage{amssymb}
\usepackage{amsmath}
\usepackage{verbatim}
\usepackage{multirow}
\usepackage{subfigure}
\usepackage{array}
\usepackage{fp}
\usepackage{makecell}
\usepackage{flushend}
\usepackage{cases}
\usepackage{color}
\usepackage{bbm}
\newcommand{\PreserveBackslash}[1]{\let\temp=\\#1\let\\=\temp}
\newcolumntype{C}[1]{>{\PreserveBackslash\centering}p{#1}}
\newcolumntype{R}[1]{>{\PreserveBackslash\raggedleft}p{#1}}
\newcolumntype{L}[1]{>{\PreserveBackslash\raggedright}p{#1}}
\newcommand{\thickhline}{\noalign{\hrule height 1pt}}

\title{Detecting Context Dependent Messages in a Conversational Environment}

	\author{
		Chaozhuo Li$^\dag$,~~~ Yu Wu$^\dag$,~~~ Wei Wu$^\ddag$,~~~ Chen Xing$^\diamond$,~~~ Zhoujun Li$^\dag$,~~~ Ming Zhou$^\ddag$~~~~\\
		$^\dag$State Key Lab of Software Development Environment, Beihang University, Beijing, China\\
		$^\ddag$~~~~Microsoft Research, Beijing, China\\
		$^\diamond$~~~~Nankai University, Tianjin, China\\
		\{lichaozhuo,wuyu,lizj\}@buaa.edu.cn \{wuwei,v-chxing,mingzhou\}@microsoft.com 
	}

\begin{document}
\maketitle
\begin{abstract}
While automatic response generation for building chatbot systems has drawn a lot of attention recently, there is limited understanding on when we need to consider the linguistic context of an input text in the generation process. The task is challenging, as messages in a conversational environment are short and informal, and evidence that can indicate a message is context dependent is scarce.
After a study of social conversation data crawled from the web, we observed that some characteristics estimated from the responses of messages are discriminative for identifying context dependent messages.
With the characteristics as weak supervision, we propose using a Long Short Term Memory (LSTM) network to learn a classifier.  Our method carries out text representation and classifier learning in a unified framework.  Experimental results show that the proposed method can significantly outperform baseline methods on accuracy of classification.
\end{abstract}

\section{Introduction}

%
%
\blfootnote{
	%
	%
	\hspace{-0.65cm}  
	This work is licensed under a Creative Commons Attribution 4.0 International Licence. Licence details: http://creativecommons.org/licenses/by/4.0/.
	%
	%
	%
	%
}

Together with the rapid growth of social media such as Twitter and Weibo, the amount of conversation data on the web has tremendously increased. This makes building open domain chatbot systems with data-driven approaches possible. To carry on reasonable conversations with humans, a chatbot system needs to generate proper response with regard to users' messages. Recently, with the large amount of conversation data available, learning a response generator from data has drawn a lot of attention \cite{ritter2011data,shang2015neural,vinyals2015neural}.
	
A key step to coherent response generation is determining when to consider linguistic context of messages. Existing work on response generation, however, has overlooked this step. They either totally ignores linguistic context \cite{ritter2011data,shang2015neural,vinyals2015neural} or simply considers context for every message \cite{sordoni2015neural,serban2015building}. The former case is easy to lead to irrelevant responses when users' input messages rely on the context information in previous conversation turns, while the latter case is costly (e.g., on memory and responding time) for building a real chatbot system and has the risk of bringing in noise to response generation especially when users want to end the current conversation topic and start a new one.  According to our observation, there are two types of messages in a conversational environment. The first type is context dependent message, which means to reply to the message, one must consider previous utterances in the dialogue\footnote{Broadly speaking, context may not be limited to linguistic context. For example, a user's interest could also be a kind of context. As the first step, in this work, we only focus on ``linguistic context''.}, while the second type is context independent message, which means even without the previous utterances, the message itself can still lead to a reasonable response. Table \ref{contextvsnoncontext} compares the two types of messages using examples. In Case 1, ``why do you think so'' is a context dependent message. In order to reply to the message, one cannot ignore its linguistic context ``I think it will rain tomorrow''. On the other hand, in Case 2, ``Well, what time is it now'' is a context independent message, as one can give a reasonable response without looking at the previous turns.  Distinguishing context dependent messages from context independent messages is important for building a good response generator. Missing linguistic context for context dependent messages will lead to nonsense response. For example, ``because I love you'' could also be a response for the message ``why do you think so'' if we only look at the message itself, but it is nonsense appearing in the dialogue of Case 1. Incorporating context information into context independent messages will increase the workload of a generation system and has the risk of bringing in noise to the generation process. For example, if we consider the context ``NBA'' for the message ``Well, what time is it now'', the chatbot will probably say something about ``NBA'' rather than answer the question with a time answer. Although detecting context dependent messages is crucial for building chatbot systems, there is limited understanding about it.
		
				\begin{table}
					\caption{Two types of messages}	
					\label{contextvsnoncontext}
					\centering
					\small
					\begin{tabular}{l|l}
						\hline
						Case 1 : a context dependent message &  Case 2 : a context independent message  \\ \hline
						User \ \ \ \ : What will the weather be like tomorrow? & User \ \ \ \ : What are you doing? \\
						Chatbot : I think it will rain tomorrow. & Chatbot : I am waiting for you to watch NBA. \\
						User \ \ \ \ : \emph{\textbf{Why do you think so?}} & User \ \ \ \ : \emph{\textbf{Well, what time is it now?}} \\
						\hline
					\end{tabular}
				\end{table}

In this paper, we study this important but less explored problem.  Instead of answering how to incorporate context information, we try to understand when we need the information. Therefore, our effort is complementary to the existing work on response generation. It can keep the existing generation algorithms context-aware and improve their efficiency and robustness to noise.  The task is challenging, as messages in a conversational environment are usually short and informal, and evidence that can indicate a message is context dependent is scarce. For example, on $3$ million post-response pairs crawled from Weibo, the average length of messages is $4.65$.  On such short texts, classic NLP tools such as POS Tagger and Parser suffer from bad performance \cite{derczynski2013twitter,foster2011hardtoparse} and it is difficult to explicitly extract features that are discriminative on the two types of messages. More seriously, there are no large scale annotations available for building a supervised learning procedure.
	
We consider leveraging the large amount of human-human conversation data available on the web to learn a message classifier. Our intuition is that a context dependent message has different linguistic context in different conversation sessions, therefore its responses could be more diverse on content than responses of a context independent message. To verify this idea, we study the distributions of responses of messages using conversation data crawled from social media and find that the length distribution of responses and the word distribution of responses are quite discriminative on the two types of messages. Based on this observation, for each message in the crawled data, we estimate the average length of responses, the entropy of the word distribution of responses, and the maximum mass of the word distribution of responses, and take these characteristics as weak supervision signals to learn a classifier. The classifier takes a message as input and can make prediction for any messages in a real conversation environment, even though the messages do not appear in the crawled data and characteristics like entropy are not available for them.  We propose using a Long Short Term Memory (LSTM) architecture to learn the classifier. Our model represents message texts in a continuous vector space using a one-layer LSTM network. The text vectors are then provided as input to a two-layer feed-forward neural network to perform classification. The neural network architecture carries out feature learning and model learning in a unified framework, and thus can avoid explicit feature extraction which is difficult on short conversational messages. Our method leverages large scale weak supervision signals extracted from responses in social conversation data and can reach a satisfactory accuracy with only a few human annotations.
	
We conduct experiments on large scale English and Chinese conversation data mined from Twitter and Weibo respectively, and test the performance of our method on thousands of messages annotated by human labelers. Experimental results show that our method can significantly outperform baseline methods on accuracy of message classification on both of the two data sets. 
	
We make the following contributions in this paper: 1) proposal of detecting context dependent messages in a conversational environment; 2) proposal of learning weak supervision signals from responses of messages using large scale conversation data; 3) proposal of using an LSTM architecture to learn a message classifier; 4) empirical verification of the proposed method on human annotated data.

\section{Related Work}

Our work lies in the path of building chatbot systems with data-driven approaches.  Differing from traditional dialogue systems (cf., \cite{young2013pomdp}) which rely on hand-crafted features and rules to generate reply sentences for specific applications such as 
voice dialling \cite{williams2008demonstration} and appointment scheduling \cite{janarthanam2011day} etc., recent effort focuses on exploiting an end-to-end approach to learn a response generator from social conversation data for open domain dialogue \cite{koshinda2015machine,higashinaka2016syntactic}. For example, Ritter et al.~\cite{ritter2011data} employed a phrase-based machine translation model for response generation. In \cite{shang2015neural,vinyals2015neural}, neural network architectures were proposed to learning response generators from one-round conversation data. Based on these work, Sordoni et al.~\cite{sordoni2015neural} incorporated linguistic context into the learning of response generator. Serban et al.~\cite{serban2015building} proposed a hierarchical neural network architecture to building context-aware response generation. In this paper, instead of studying how to incorporate context into response generation, we consider the problem that when we need context in the process. Our work can keep the existing generation algorithms context-aware and at the same time improve their efficiency and robustness.
	
	We employ a Recurrent Neural Network (RNN) architecture to learn a message classifier. RNN models \cite{elman1990finding}, due to their capability of modeling sequences with arbitrary length, have been widely used in many natural language processing tasks such as language modeling \cite{mikolov2010recurrent} and tagging \cite{xu2015ccg} etc. Recently, it is reported that Long Short Term Memory (LSTM) \cite{hochreiter1997long} and Gated Recurrent Unit (GRU) \cite{cho2014learning} as two special RNN models which can capture long term dependencies in sequences outperform state of the art methods on tasks like machine translation \cite{sutskever2014sequence} and response generation \cite{shang2015neural}. In this paper, we apply the LSTM architecture to the task of context dependent message detection. We append LSTM with a two-layer feed-forward neural network, thus feature learning and model learning can be carried out simultaneously.
	
	Our work belongs to the scope of short text classification \cite{song2014short}.  Existing applications of short text classification include query classification \cite{kang2003query}, tweet classification \cite{sriram2010short}, and question classification \cite{zhang2003question}. We study a new problem in short text classification: distinguishing context dependent messages from context independent messages in a conversational environment. The task is important for building open domain chatbot systems and has its unique challenges (e.g., new data structure). We tackle the challenges by leveraging the responses of messages and utilizing an LSTM network to conduct feature learning and model learning simultaneously.

	\section{Learning to Detect Context Dependent Messages}
	Suppose that we have a data set $\mathcal {D}=\{(m_i, y_i)\}_{i=1}^N$ where $m_i$ is a message composed of a sequence of words $(w_{m_i,1},\ldots, w_{m_i,n_i})$ and $y_i$ is an indicator whose value reflects whether $m_i$ is context dependent or not. Our goal is to learn a function $g(\cdot) \in \{-1,1\}$ using $\mathcal{D}$, thus for any new message $m$, $g(\cdot)$ predicts $m$ a context dependent message if $g(m)=1$. To this end, we need to answer two questions: 1) how to construct $\mathcal{D}$; 2) how to perform learning using $\mathcal{D}$.

	For the first question, we can crawl conversation data from social media like Twitter and ask human labelers to annotate the messages in the data.  The problem is that human annotation is expensive and time consuming and therefore we cannot obtain a large scale data set for learning. To solve the problem, we automatically learn some weak supervision signals using responses of messages in social conversation data, and take the signals as $\{y_i\}$ in $\mathcal{D}$. For the second question, one straightforward way is first extracting shallow features such as bag-of-words and syntax from messages and then employing off-the-shelf machine learning tools to learn a model. The problem is that shallow features are not effective enough on representing semantics in short conversation messages, which will be seen in our experiments. We propose using a Long Short Term Memory (LSTM) architecture to learn a model from $\mathcal{D}$. The advantage of our approach is that it can avoid explicit feature extraction and large scale human annotations, and carry out feature learning and model learning in a unified framework.

	\subsection{Learning Weak Supervision Using Responses}\label{weaksuper}

	Instead of requiring human annotations, we consider creating signals that are discriminative on the two types of messages from large scale social conversation data available on the web. Our intuition is that a context dependent message has different linguistic context in different conversation sessions, therefore, its responses could be more diverse on content than responses of a context independent message (one message may appear multiple times, and therefore it may correspond to multiple responses).  Table 2 illustrates our idea with some examples from Twitter. The last column of the table represents the frequency of the message or the frequency of the response under the message. For each message, we show the top 5 most frequent responses. From the examples, we can see that a context dependent message tends to have divergent and uniformly distributed responses corresponding to different linguistic context, while the responses of a context independent message share relatively similar content and some content dominates the distribution.
	
			\begin{table}[h]
				\label{example}
				\small
				\caption{Responses of the two types of messages}
				\centering
				\begin{tabular}{p{19em}|c|p{19em}|c}
					\hline
					Context dependent message : why& 2196 & Context independent message : Good night & 644 \\ \hline
					Response 1 : I am kidding & 7 & Response 1 : Good night & 47 \\
					Response 2 : He can be like mcdaniels for sixer & 5 & Response 2 :  Goodnight & 44\\
					Response 3 : Because I say no & 5 & Response 3 : Night & 23 \\
					Response 4 : I am tired & 5 & Response 4 : Sleep well & 10  \\
					Response 5 : U will become dependent on them & 5 & Response 5 : Thank you & 9\\
					\hline
				\end{tabular}
			\end{table}
			
	The examples inspire us to investigate some statistical characteristics that can reflect the diversity of responses. These characteristics could be good indicators of context dependent messages, and we can construct $\{y_i\}$ in $\mathcal{D}$ using the characteristics. We estimate the following statistical characteristics for each message using its responses, and examine how the characteristics are discriminative on the two types of messages using $1000$ labeled messages from Twitter and Weibo respectively. The details of the labeled data will be described in our experiments.

\begin{figure*}
				\centering
				\subfigure[Comparison on entropy]{
					\label{fig:entropy} 
					\includegraphics[width=5.0cm,height=4.1cm]{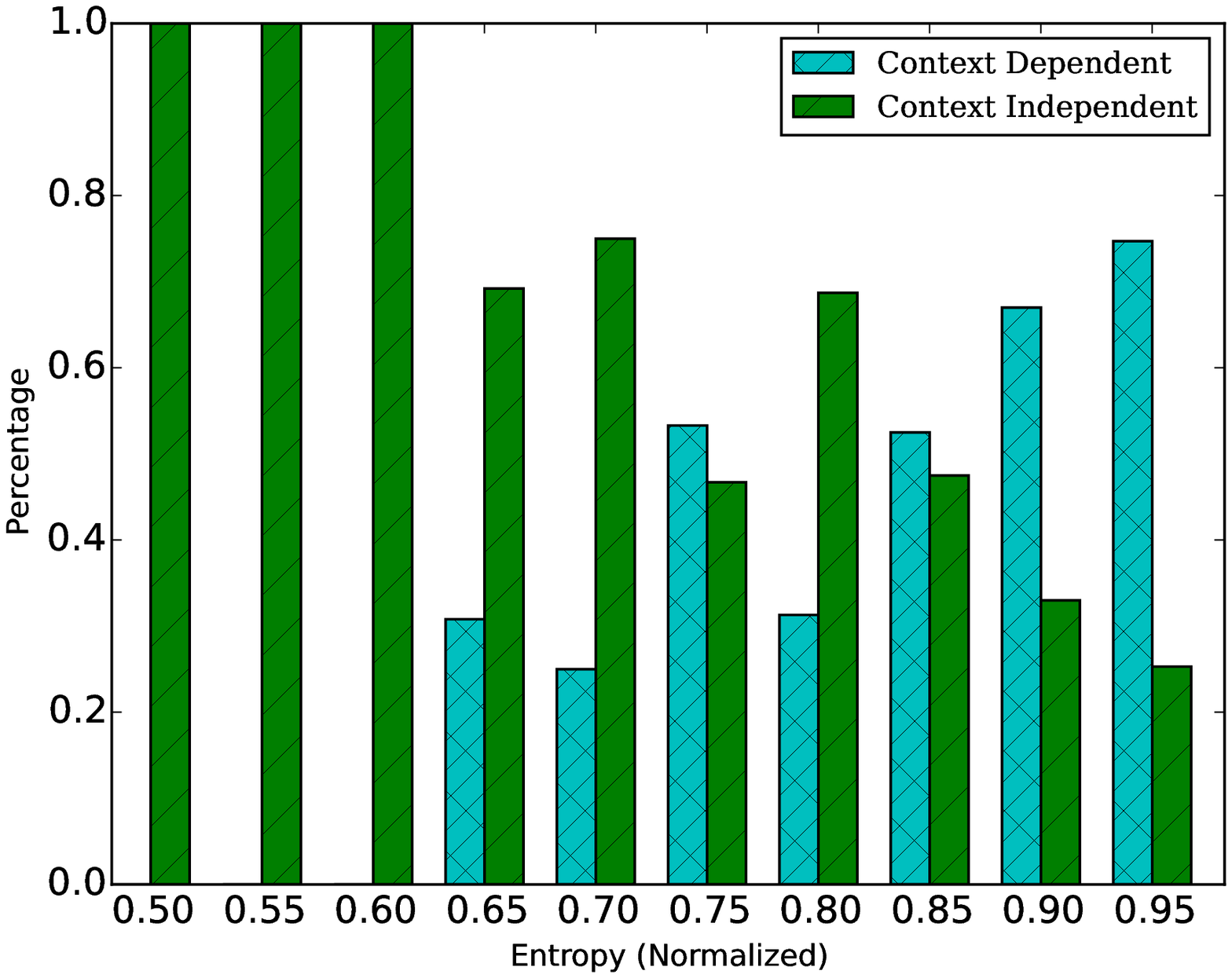}}
				\subfigure[Comparison on $M(P)$]{
					\label{fig:maxmass} 
					\includegraphics[width=5.0cm,height=4.1cm]{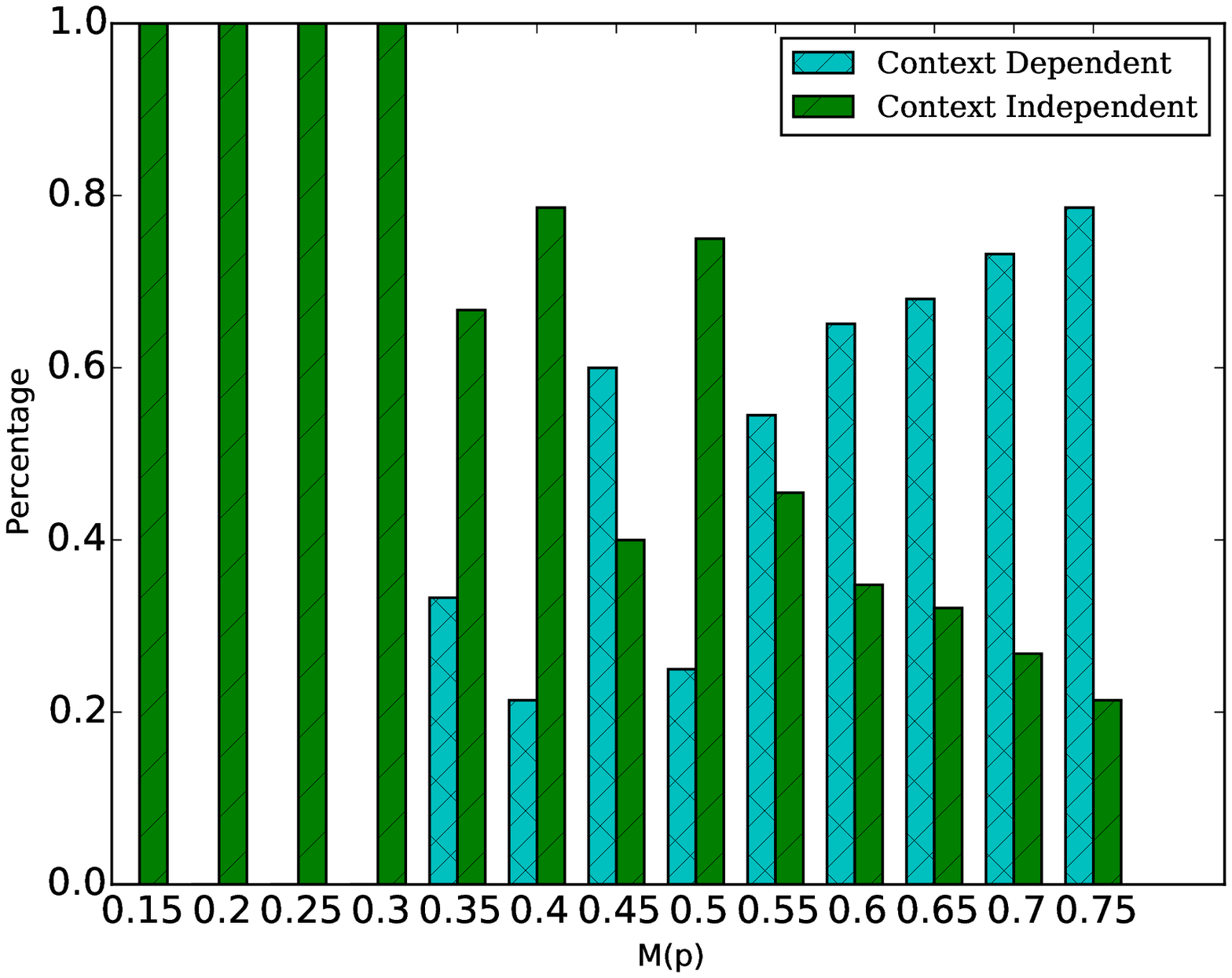}}
				\subfigure[Comparison on average length of responses]{
					\label{fig:avelength} 
					\includegraphics[width=5.0cm,height=4.1cm]{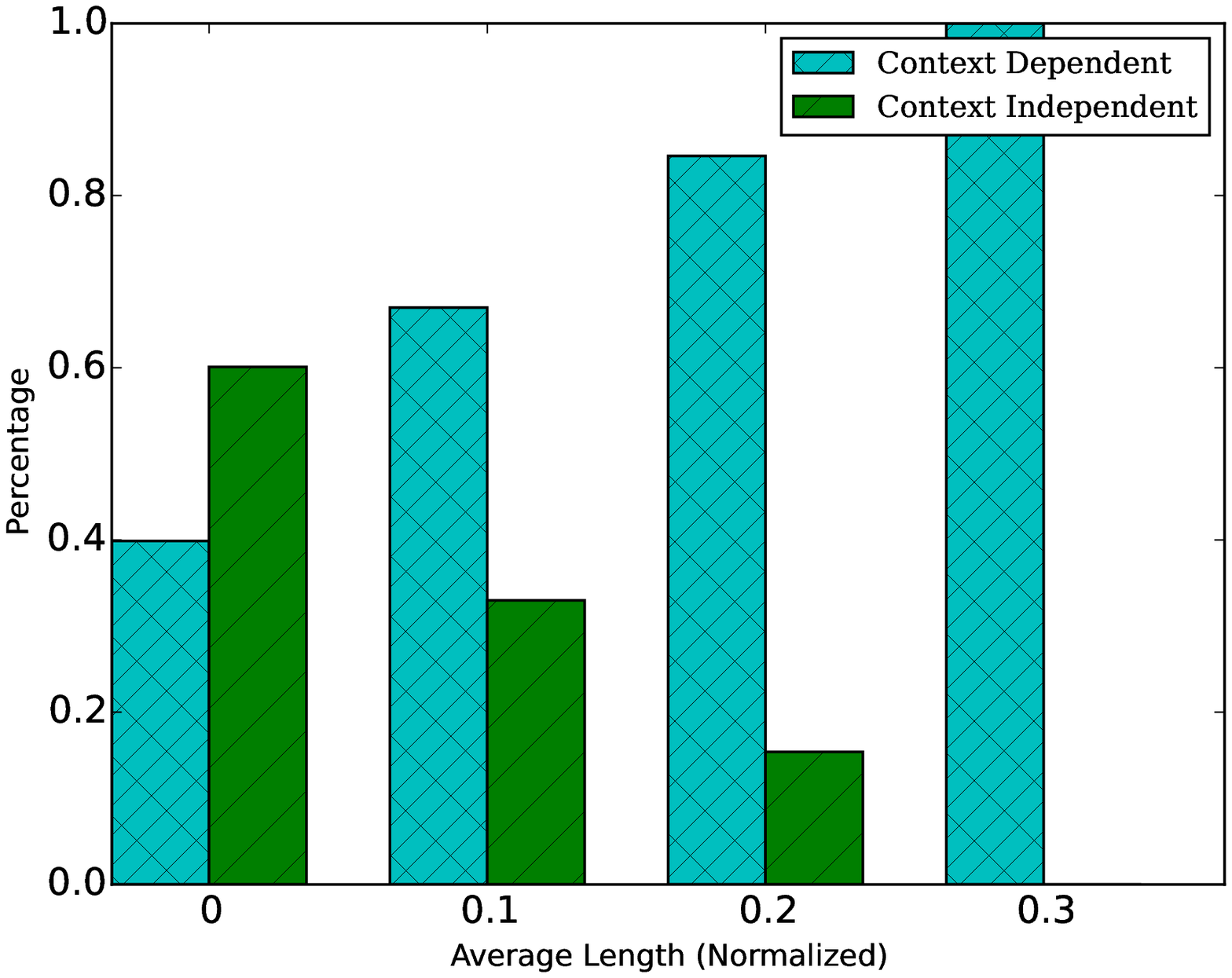}}
				\caption{Comparison of the two types of messages on three characteristics.}
				\label{fig:compareall} 
			\end{figure*}

	\textbf{Entropy:} the first characteristic we investigate is the entropy of the word distribution of responses, which is a common measure for diversity. Given a word distribution $P=(p_1, p_2, \ldots, p_n)$, the entropy of the distribution is defined as
	\begin{equation}\label{entropy}
	\small
	E(P)=\sum_{i=1}^n -p_i \log_2(p_i).
	\end{equation}
	The maximum of the entropy is $\log_2(n)$ which is reached when the distribution is uniform. Then, a large entropy means a word distribution covers many words (i.e., $n$ is big) and is close to a uniform distribution. Therefore, a context dependent message should have a larger entropy on responses than a context independent message (see the comparison in Table 2). We normalize the entropy to $[0,1]$ by $\frac{E(P)-\min(E)}{\max(E)-\min(E)}$, where $\max(E)$ and $\min(E)$ represent the maximum entropy and the minimum entropy in the data set.  Figure \ref{fig:entropy} shows the comparison of the two types of messages on normalized entropy using the Twitter labeled data. In the figure, each value on the x-axis represents an interval with a fixed length $0.05$. For example, $0.50$ means an interval $[0.5,0.55)$. Each value on the y-axis represents the percentage of messages in a specific interval. For example, among messages falling in the interval $[0.95, 1)$, nearly $80\%$ are labeled as context dependent and only about $20\%$ are labeled as context independent. From the figure, we can see that entropy is discriminative on the two types of messages: context dependent messages distributes on large entropy areas, while context independent messages tend to have smaller entropy.

	\textbf{M(P)}: in addition to entropy, another characteristic that might reflect the diversity of responses could be the maximum mass of the word distribution of responses, as in diverse responses, words should be uniformly distributed (stopwords are removed), while in less diverse responses, there may exit dominant words (e.g., ``night'' in Table 2). Given a word distribution $P=(p_1, p_2, \ldots, p_n)$, we define a characteristic as
	\begin{equation}\label{maxmass}
	\small
	M(P) = 1 - \underset{1 \leqslant i \leqslant n}{\max} p_i
	\end{equation}
	Figure \ref{fig:maxmass} compares the two types of messages on $M(P)$ using the Twitter labeled data, in which values on the x-axis and y-axis have the same meaning as those in Figure \ref{fig:entropy}. From the figure, we can see that similar to entropy, $M(P)$ is useful on distinguishing the two types of messages. Context dependent messages have larger $M(P)$ than context independent messages.

	\textbf{Average length of responses}: finally, we consider the length distribution of responses. Since responses of context dependent messages are more diverse on content, they might be longer than responses of context independent messages. We calculate the average length of responses for each message and normalize it to $[0,1]$ in the same way as entropy. Figure \ref{fig:avelength} compares the two types of messages on average length of responses using the Twitter labeled data, where values on the x-axis represent intervals with a length $0.1$. The result supports our claim and clearly indicates that average length is discriminative on the two types of messages.

	We combine the three characteristics using a linear SVM classifier learned with the $1000$ labeled messages and take the output of the SVM (a real value) as $\{y_i\}$ in $\mathcal{D}$. By this means, we can create a large scale training data set with only a little human labeling effort. Here, as a reference, we also report the classification accuracy of the three characteristics and the SVM classifier on the $1000$ labeled data. Each characteristic corresponds to a threshold tuned on the $1000$ labeled data with $5$-fold cross validation. If a value of a characteristic of a message is larger than the threshold, then the message will be predicted as context dependent.  Table \ref{3chara} shows the classification accuracy of $5$-fold cross validation (average of $5$ results), where SVM (com) refers to the SVM classifier. Details of experiment setting will be described in Section \ref{exp}. From Table \ref{3chara}, we can see that the numbers are consistent with Figure \ref{fig:entropy}, \ref{fig:maxmass}, and \ref{fig:avelength}.
\begin{table}[h]			
\centering
\small
			\caption{Classification accuracy on $1000$ labeled data}
\label{3chara}			
\begin{tabular}{l|c|c}
				\hline
				& Weibo & Twitter \\ \hline
				Entropy & 72.6 \% & 70.5 \% \\ \hline
				$M(P)$ & 72.6 \% &  69.8 \% \\ \hline
				Average length of responses & 72.8 \% & 68.5 \% \\ \hline
				SVM (com) & 73.8 \% & 71.2 \% \\ \hline
			\end{tabular}
	\end{table}
\subsection{Model Learning}\label{ML}	
We head for learning $g(\cdot)$ using $\mathcal{D}$ constructed in Section \ref{weaksuper}. Note that $g(\cdot)$ only takes a message $m$ as input, and thus can make prediction for any messages in a real chatbot system even though the messages are not in $\mathcal{D}$ and their entropy, M(P), and average length of responses are not available. Our idea is that we first learn a regression model by fitting $\{y_i\}$ in $\mathcal{D}$ through minimizing the sum of squared residuals and then construct $g(\cdot)$ by comparing the output of the regression model with a threshold. We can obtain the threshold by tuning it on a few labeled data (e.g., the $1000$ labeled data). The key is how to learn the regression model. We propose using a Recurrent Neural Network (RNN) architecture to embed messages into a continuous vector space and learning a regression model with the embedding of messages using a feed-forward neural network. The RNN model, which is capable of embedding sequences with arbitrary length, can encode the order of words and the semantics of a message into a vector representation which has been recently proven effective on capturing similarity of short texts \cite{sordoni2015hierarchical}. We take the output vector given by RNN as a feature representation of a message and feed it to a feed-forward neural work. By this means, we can conduct feature learning and model learning in a unified framework and jointly optimize the two components.	
					\begin{figure*}[h]		
						\begin{center}
							\includegraphics[width=16cm,height=3.67cm]{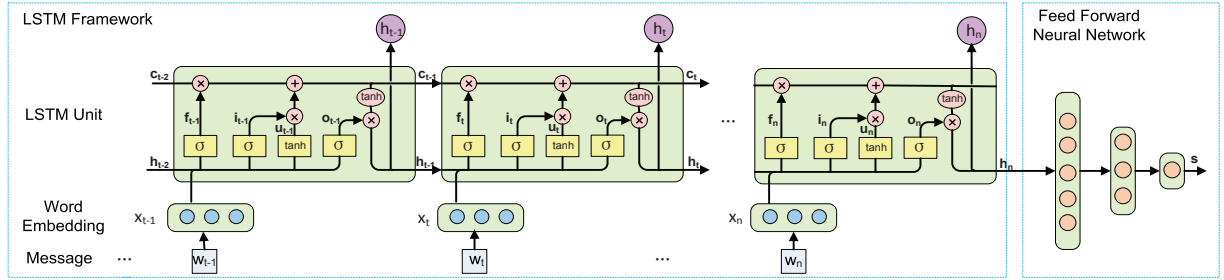}
						\end{center}
						\caption{The architecture of our method}\label{fig:framework}
					\end{figure*}

Given a message $m$ which consists of $n$ words, the RNN model reads the words one by one, and updates a recurrent state $h_t$ for the $t$-th word $w_t$ by
	\begin{equation}
	\small
	h_t=f(h_{t-1}, x_t), \thickspace h_0=0,
	\end{equation}
	where $h_t \in \mathbb{R}^{d_h}$, $x_t \in \mathbb{R}^{d_w}$ is the vector representation of $w_t$, and $f$ is non-linear transformation. $h_t$ acts as an encoding of the semantics of the word sequence up to position $t$, and the final output $h_n$ is a representation of message $m$.  Both $x_t$ and $h_t$ are learned in the optimization of the RNN model. We select the Long Short Term Memory (LSTM) \cite{hochreiter1997long} as $f$, since it can model long term dependencies in sequences with affordable complexity. LSTM controls the learning of the representation of a sequence by gates. Specifically, at position $t$, LSTM controls the information that should be kept from previous states by an input gate $i_t$, and the information that should be forgotten by a forget gate $f_t$. After memorizing and forgetting, the information is stored in a memory cell $c_t$. $c_t$ generates the recurrent state $h_t$ through an output gate $o_t$. The specific parameterization of LSTM is given by
	\begin{equation*}
	\small
	\begin{split}
	i_t& = \sigma(W^{(i)}x_t + U^{(i)}h_{t-1}+b^{(i)}) \\
	f_t& = \sigma(W^{(f)}x_t + U^{(f)}h_{t-1}+b^{(f)}) \\
	o_t& = \sigma(W^{(o)}x_t + U^{(o)}h_{t-1}+b^{(o)}) \\
	u_t& = \tanh(W^{(u)}x_t + U^{(u)}h_{t-1}+b^{(u)}) \\
	c_t& =  i_t \otimes u_t + f_t \otimes c_{(t-1)}\\
	h_t& = o_t \otimes \tanh(c_t),
	\end{split}
	\end{equation*}
	where $\sigma(\cdot)$ is a sigmoid function and $\tanh(\cdot)$ is a hyperbolic tangent function. $W^{(i)}$, $W^{(f)}$, $W^{(o)}$, $W^{(u)}$ $\in \mathbb{R}^{d_h\times d_w}$, $U^{(i)}$, $U^{(f)}$, $U^{(o)}$, $U^{(u)}$ $\in  \mathbb{R}^{d_h\times d_h}$, and $b^{(i)}$, $b^{(f)}$, $b^{(o)}$, $b^{(u)}$ $\in \mathbb{R}^{d_h \times 1}$ are parameters. $\otimes$ means element-wise multiplication. After we get the final state $h_n$, we feed it to a two-layer feed-forward neural network to get an output $s$ which is defined by
	\begin{equation}\label{eq:mlp}
	\small
	s = b_2 + W_2 \left(\tanh(b_1 + W_1 h_n)\right),
	\end{equation}
	where $b_1 \in \mathbb{R}^{d_s \times 1}$, $W_1 \in \mathbb{R}^{d_s \times d_h}$, $W_2 \in \mathbb{R}^{1 \times d_s}$, and $b_2 \in \mathbb{R}$ are parameters.  Figure \ref{fig:framework} illustrates the architecture of our method.

	For each $m_i$ in $\mathcal{D}$, we calculate an $s_i$ using Equation (\ref{eq:mlp}) as an estimation of $y_i$. We then learn the parameters of the LSTM network and the feed-forward network by minimizing the sum of the squared residuals. Formally, our learning approach can be formulated as
	\begin{equation}\label{eq:learning}
	\small
	\underset{s}{\arg\min} \thickspace \sum_{i=1}^N (y_i-s_i)^2.
	\end{equation}
	After we obtain the parameters, we can calculate an $s_m$ for any message $m$ using Equation (\ref{eq:mlp}). We then tune a threshold $T$ with a few labeled messages. The classifier $g(\cdot)$ is given by
	\begin{equation}\label{eq:predict}
	\small
	g(m)=\left\{
	\begin{array}{ll}
	1 & \textrm{if $s_m > T$}\\
	-1 & \textrm{otherwise}\\
	\end{array}
	\right.
	\end{equation}
	The gradients of the objective function (\ref{eq:learning}) are computed using the back-propagation through time (BPTT) algorithm \cite{williams1990efficient}. We share the code for model learning at \url{https://github.com/whatsname1991/coling2016}. 	
\section{Experiments}\label{exp}
\subsection{Experiment Setup}
We constructed the conversation data for experiments from Weibo and Twitter. In each of the two social media, two persons can communicate by replying to each other under a post. We crawled sequences of reply with posts and  extracted triples like ``(context, message, response)'' as experimental data. In a triple, ``message'' is a reply, ``context'' is the sentence in the previous turn of the message (a reply or a post), and ``response'' is the sentence in the next turn (reply to the message). Note that in this work, we restrict the context of a message to a single sentence. This is a simplification of context in conversation. In real conversation, context could be more complicated and we leave the discussion of it as future work.

We crawled $5.9$ million English triples from Twitter, and $3.1$ million Chinese triples from Weibo. The numbers of distinct messages in the Twitter data and in the Weibo data are $92,755$ and $112,175$ respectively. On average, each Twitter message has $63.26$ responses (some messages like ``hello'' can have many different responses) and each Weibo message has $27.52$ responses. The average word length of Twitter message is $3.39$ and the word average length of Weibo message is $4.65$. English sentences were stemmed and stop words were removed, and Chinese sentences were segmented.
	
We constructed $\mathcal {D}=\{(m_i, y_i)\}_{i=1}^N$ in Section \ref{weaksuper} in the following way: we first calculated entropy, $M(P)$, and average length of responses for each message using the $5.9$ million English triples and $3.1$ million Chinese triples. Then from these data, we randomly sampled $1000$ English triples and $1000$ Chinese triples as validation sets. For each triple in the validation data, we hid the response and recruited human judges to label if the message is context dependent or not. Note that we hid responses when labeling messages because this is more close to the real case. In a real chatbot system, one has to determine if a message is context dependent or not before generating a response.  Each judge labeled a message with $1$ if it is context dependent, otherwise the judge labeled the message with $-1$. Each message got three labels and the majority of the labels was taken as the final decision for the message.  In the Weibo data, there are $412$ positive examples and $588$ negative examples. In the Twitter data, the two numbers are $440$ and $560$, respectively. With the two validation data sets, we learned two SVM classifiers in order to combine the three characteristics as described in Section \ref{weaksuper}. Parameters of SVMs were tuned by $5$-fold cross validation. Finally, we assigned a $y_i$ to each $m_i$ in the $112,175$ Twitter messages and $92,755$ Weibo messages by the output of the SVM classifiers, and formed $\mathcal{D}$ for both English data and Chinese data. We trained LSTM models using $\mathcal{D}$.

To evaluate the performance of different models, we crawled another $3000$ Chinese context-message pairs and $1000$ English context-message pairs from Weibo and Twitter respectively, and followed the same way as the validation data to judge if the messages are context dependent or not. We used these data to simulate real context-message pairs in chatbot systems. In the Weibo data, there are $2715$ unique messages and $1983$ messages are not in $\mathcal{D}$. The numbers of positive examples and negative examples are $1472$ and $1528$ respectively.  In the Twitter data, the number of unique messages is $875$ and $366$ messages are not included by $\mathcal{D}$. The numbers of positive and negative examples are $464$ and $536$ respectively. Note that for messages that are not included by $\mathcal{D}$, their characteristics (i.e., entropy, $M(P)$, and average length of responses) are not available, and we can only use classifiers whose features are extracted from messages (like our LSTM models) to make prediction. This is close to a real situation in chatbots, and we took the two data sets as test sets.


	We considered the following methods as baselines:
	
	\textbf{Length}: intuitively, short messages tend to be context dependent (e.g., ``why'' in Table 2). Therefore, we employed length of a message as a baseline. A message shorter than a threshold will be predicted as a context dependent message. 
	
	\textbf{MDF}: given a word, we estimated the number of messages that contain the word  and named it ``document frequency'' (DF). We constructed a list of words associated with DF using $\mathcal {D}$. For a new message, we calculated the minimal DF of words in the message using the list. A context dependent message like ``why do you think so'' may consist of common words, and thus correspond to a high minimal DF.  We considered minimal DF as a baseline. A message with a minimal DF larger than a threshold will be predicted as a context dependent message. 
	
	\textbf{SVM (Length+MDF)}: we linearly combined Length and MDF by learning an SVM classifier on the validation data. 
	
	\textbf{SVM (classification)}: we extracted unigrams, bigrams, and frequencies of POS tags as features from a message, and learned a linear SVM classifier on the validation data with these features. POS tags for Chinese data were obtained using Stanford Parser (\url{http://nlp.stanford.edu/software/lex-parser.shtml}) and POS tags for English data were obtained using TweetNLP (\url{http://www.cs.cmu.edu/~ark/TweetNLP/}). 
	
	\textbf{SVM (regression)}: instead of learning a classifier from annotations in the validation data, we fitted $\{y_i\}$ in $\mathcal{D}$ by learning an SVM regression model using the same features as SVM (classification) and made predictions on new messages by a threshold. 
	
All SVM models were learned using SVM-Light (\url{http://svmlight.joachims.org/}). We employed classification accuracy as an evaluation metric.
	\subsection{Parameter Tuning}
	For Length and MDF, the only parameter is a threshold. We tuned the thresholds on the validation data. For all SVM models, we selected the trade-off parameter in SVM from $\{0.01, 0.1, 1, 10, 100\}$ by $5$-fold cross validation on the validation data. SVM (regression) also needs a threshold. We tuned it on the validation data. The parameters of LSTM include the dimension of word vectors $d_w$, the dimension of hidden states $d_h$, and the dimension of the first layer of the feed-forward network $d_s$. We set $d_w=d_h=256$, and $d_s=100$. Besides these parameters, we also set a dropout rate $0.1$ in the learning of the feed-forward network as regularization.
				 		\begin{table}[h]
				 			\begin{minipage}{.5\linewidth}
				 				\centering
				 				\caption{Accuracy on two test sets}
				 				\label{experimentResults}
				 				\begin{tabular}{l|c|c}
				 					\thickhline
				 					& Weibo & Twitter \\ \hline
				 					Length & 62.6 \% & 61.3 \% \\ \hline
				 					MDF & 62.1 \% &   58.6 \% \\ \hline
				 					SVM (Length+MDF)  & 63.0 \% & 62.2 \% \\ \hline
				 					SVM (classification) & 66.8 \% & 65.4 \% \\ \hline
				 					SVM (regression) & 64.3 \% & 68.3 \% \\ \hline
				 					LSTM & \textbf{75.6 \%} & \textbf{73.4 \%}\\ \thickhline
				 				\end{tabular}
				 			\end{minipage}\begin{minipage}{.5\linewidth}  
				 			\small
				 			\caption{Comparison between LSTM,  SVM (classification), and SVM (regression)}
				 			
				 			\label{tab:case2}
				 			\begin{tabular}{l|p{11.5em}}
				 				\thickhline
				 				Example & context : Have you heard Taylor Swift's new song? \\
				 				& \emph{message:  Yep, I have heard it on Saturday night.} \\
				 				Label & context dependent \\ \hline
				 				SVM (regression) & context independent\\ \hline
				 				SVM (classification) & context independent \\ \hline
				 				LSTM  & context dependent \\
				 				\thickhline
				 			\end{tabular}
				 		\end{minipage}
				 	\end{table}
				 	
	\subsection{Quantitative Evaluation}
	Table \ref{experimentResults} reports quantitative evaluation results on the test data.  From the results, we can see that our methods outperform baseline methods. The improvement over the best performing baseline methods (i.e., SVM (classification) on Webio and SVM(regression) on Twitter) is statistically significant (sign test, $p$-value $<0.01$).

	Length and MDF are characteristics of messages. The results tell us that these characteristics are not so discriminative on the two types of messages. The reason is easy to understand: we may think that context dependent messages tend to be short and consist of common words, but the fact is that short messages composed of common words could be context independent (e.g., ``Good night'' in Table 2) while long messages like ``Yep, I have heard it on Saturday night'' (see the example in Table 5) could be context dependent. Both SVM (classification) and SVM (regression) perform worse than our LSTM model, indicating that shallow features are not effective enough to represent the semantics in short conversation messages. Our method outperforms the baseline methods on both data sets. The results verified our idea on leveraging responses for context dependent message detection,  and demonstrates the power of big data and the advantage of LSTM on capturing semantics in short messages.	
	\subsection{Qualitative Evaluation}
	We use an example to further explain why our method is effective on distinguishing the two types of messages. Table \ref{tab:case2} compares LSTM with SVM (classification) and SVM (regression). Both SVM (classification) and SVM (regression) rely on shallow features such as bag of words and pos tags to perform learning. These features, however, are not effective on representing the semantics of short messages. The representation is easily to be biased by some specific words like ``Saturday night'' in the example. Therefore, both SVM (classification) and SVM (regression) failed on this case. On the other hand, LSTM models term dependencies in sequences with a memorizing-forgetting mechanism. It can capture the semantics in the message ``Yep, I have heard it on Saturday night.'' and identify that it is similar to messages like ``Yes, I did'' and ``Yes, I have''. For example, the cosine of the vector of ``Yep, I have heard it on Saturday night.'' and the vector of ``Yes, I have'' given by LSTM is 0.63. Since messages like ``Yes, I did'' and ``Yes, I have'' are common context dependent messages, LSTM can successfully recognize that the message in the example is also context dependent.
%
%

	\section{Conclusion}
	We propose learning a LSTM network with weak supervision signals estimated from responses of messages to detecting context dependent messages in a conversational environment. Evaluation results show that the proposed method can significantly outperform baseline methods on distinguishing the two types of messages.

	\section*{Acknowledgement}
	This work was supported by Beijing Advanced Innovation Center
	for Imaging Technology (No.BAICIT-2016001), the National Natural Science Foundation
	of China (Grand Nos. 61370126, 61672081), National High Technology Research
	and Development Program of China (No.2015AA016004),the Fund of the State Key
	Laboratory of Software Development Environment (No.SKLSDE-2015ZX-16).

\bibliographystyle{acl}
\bibliography{coling2016}

\end{document}